# Brain Tumor Segmentation Using Deep Learning by Type Specific Sorting of Images

Zahra Sobhaninia, Safiyeh Rezaei, Alireza Noroozi, Mehdi Ahmadi, Hamidreza Zarrabi, Nader Karimi, Ali Emami, Shadrokh Samavi

Department of Electrical and Computer Engineering, Isfahan University of Technology, Isfahan

*Abstract*- **Recently deep learning has been playing a major role in the field of computer vision. One of its applications is the reduction of human judgment in the diagnosis of diseases. Especially, brain tumor diagnosis requires high accuracy, where minute errors in judgment may lead to disaster. For this reason, brain tumor segmentation is an important challenge for medical purposes. Currently several methods exist for tumor segmentation but they all lack high accuracy. Here we present a solution for brain tumor segmenting by using deep learning. In this work, we studied different angles of brain MR images and applied different networks for segmentation. The effect of using separate networks for segmentation of MR images is evaluated by comparing the results with a single network. Experimental evaluations of the networks show that Dice score of 0.73 is achieved for a single network and 0.79 in obtained for multiple networks.**

## I. INTRODUCTION

Brain tumors are the consequence of abnormal growths and uncontrolled cells division in the brain. They can lead to death if they are not detected early and accurately. Some types of brain tumor such as Meningioma, Glioma, and Pituitary tumors are more common than the others.

Meningiomas are the most common type of tumors that originate in the thin membranes that surround the brain and spinal cord. Meningiomas tumors are usually benign. The Gliomas are a collection of tumors that grow within the substance of the brain and often mix with normal brain tissue [1]. Gliomas tumors lead to a very short life expectancy when the size of the tumor is relatively large.

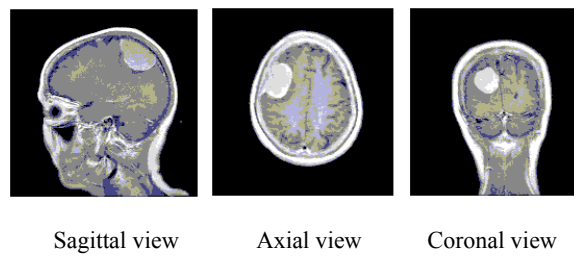

Sagittal view  Axial view  Coronal view

Figure 1: Brain MRI slices captured from different directions.

Pituitary tumors are abnormal growth of the brain cells. Pituitary tumors usually develop in the pituitary gland of the brain. Some pituitary tumors result in the abnormal and dangerous increase in the hormones that regulate important functions of the body. These tumors can appear anywhere from the brain because of their intrinsic nature. Also, they do not have a uniform shape. They have different sizes, shapes, and contrasts.

Magnetic Resonance Imaging (MRI) is a medical imaging technique, which is extensively used for diagnosis and treatment of brain tumors in clinical practice. The MR images are taken from three different directions. These views are called sagittal, axial and coronal. These three types of brain MR images are shown in Figure 1. Brain Tumor segmentation techniques are a critical component in tumor detection. Using machine learning techniques that learn the pattern of brain tumor is useful because manual segmentation is time-consuming and being susceptible to human errors or mistakes.

In general medical image segmentation is the process of automatic or semi-automatic detection of boundaries within a 2D or 3D image. In recent years many works have been done for segmentation of medical images, such as skin lesion [2], brain tumor detection [1], monitoring heart ventricles [3], and liver diagnosis [4].

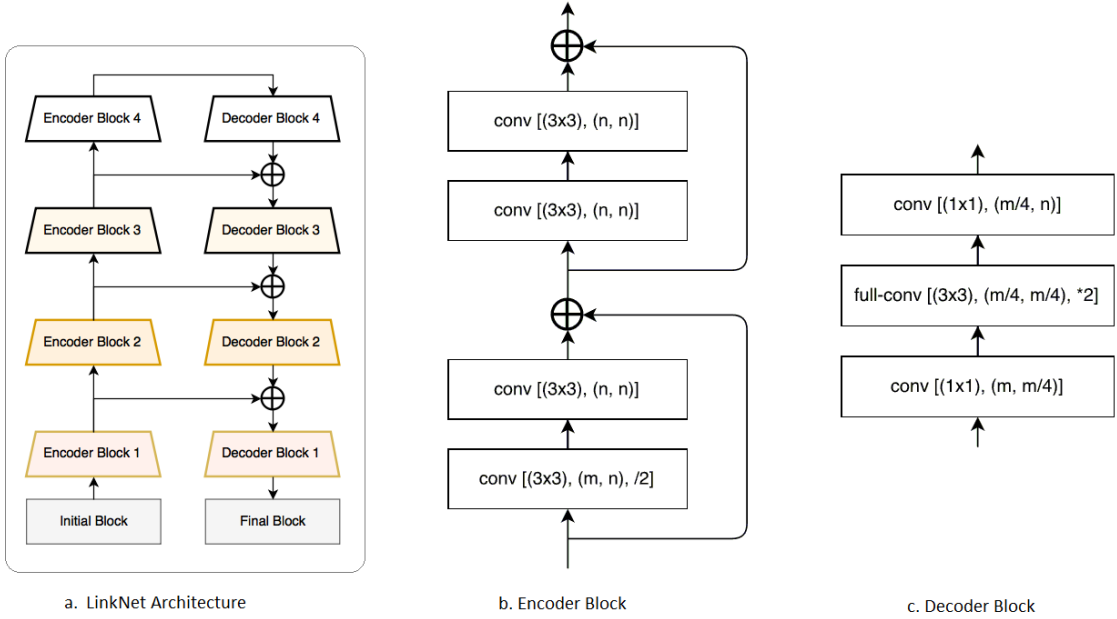

Figure 2. (a) LinkNet architecture, (b) convolutional modules in encoder-block, (c) convolutional modules in decoder-block [9]

Usually, healthy brain tissue consists of three parts: gray matter, white matter, and cerebrospinal fluid. The segmentation is used to identify areas surrounded by a tumor. The segmentation should separate the active tumorous tissue from the necrotic tissue, and also the edema (swelling near the tumor) should be identified. This is done by identifying abnormal areas when compared to normal tissue [5] [6] [7].

Most automatic brain tumor segmentation methods use hand-crafted features such as edges, corners, histogram of gradient, local binary pattern, etc. [8]. In these methods, the focus has been on implementation of a classical machine learning pipeline. The intended features are first extracted and then given to a classifier. The training procedure of the classifier is not affected by the nature of those features [5]. Convolutional neural networks (CNNs) do not use hand-crafted features and they have been applied successfully to segmentation problems.

In this work, we present an automatic brain tumor segmentation technique based on Convolutional Neural Network. We have used three MRI views of human brain. MRI scan is used because it is less harmful and more accurate than CT brain scan. All previous works on the dataset that we are working with are for classification of tumor types. None of the previous works performed on this dataset are intended for segmentation. The main contribution of our paper is the partitioning of the images based on the direction of captured MR images. Hence, three networks are trained separately to achieve better segmentation results.

## II. CONVOLUTIONAL NEURAL NETWORK

Convolutional Neural Network (CNN) is used for learning how to segment images. CNN extracts features directly from pixel images with minimal preprocessing. The network we use is LinkNet. It is a light deep neural network architecture designed for performing semantic segmentation. This network is 10 times faster than SegNet and more accurate [9].

The LinkNet Network consists of encoder and decoder blocks that arrange to break down the image and build it back up before passing it through a few final convolutional layers.

The architecture of LinkNet is presented in Fig. 2(a). The left part of the network is the encoder while the right part is the decoder. The network starts with the Initial block that implements convolution function with 7×7 kernel size and max-pooling with stride 2.

By adding the output of the encoder to the decoder, the performance of LinkNet increases because this helps the decoder to better recover the information details of the encoder-block layers. The encoder part of the network is shown in Fig. 2(b), and the layer details of the decoder-blocks are shown in Fig. 2(c) [9].

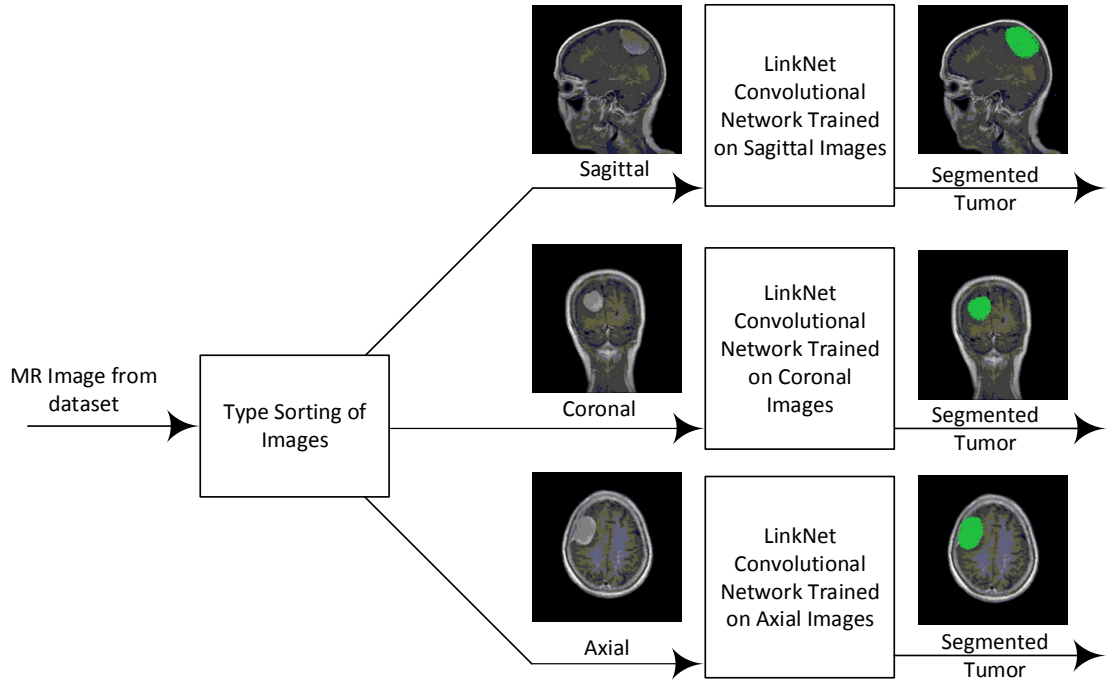

Figure 3: Overview of the proposed method.

### III. PROPOSED METHOD

In this paper, we applied a linkNet network for tumor segmentation. At first, we did not consider the view angle of the images. We initially used a single Linknet network and sent all training dataset to that network for segmentation.

All dataset images are grayscale and the foreground of the images are located at the center. Images are captured from different views of the skull; hence the size and position of the tumors vary in different angles. These differences in the size of the tumors make the diagnosis of the tumor hard. In practice, the expert physician knows the direction that the MR image is captured. Since the learning process in deep networks is similar to the human learning process, we decided to create the same situation for the deep neural networks. We found out using a single network for identification of tumors in all images does not produce accurate results.

We considered the difference network to be trained on separate MR images according to their angles. Hence, sagittal, coronal and axial images are sorted and each group is used to train one of the three networks. We used an individual LinkNet network for each of the three mentioned groups of images. Figure 3 shows our proposed method. In the next section we will show the difference in the accuracy of using only one network in contrast to the use of three separate networks.

This brain tumor T1-weighted CE-MRI image-dataset consists of 3064 slices. There are 1047 coronal images. Coronal images are captured from the back of the head. Axial images that are taken from above are 990 images. Also, there are 1027 sagittal images that are captured from the side of the skull. This dataset has a label for each image, identifying the type of the tumor. These 3064 images belong to 233 patients. The dataset includes 708 Meningiomas, 1426 Gliomas, and 930 Pituitary tumors, which are publicly available in: (http://dx.doi.org/10.6084/m9.figshare.1512427).

The network training process and details are mentioned in the followings.

For the single LinkNet network, we used 2100 of images for network training that 20% of these images are considered as validation and the rest of the data is used for the test purpose.

Also for the training of the three LinkNet networks, we separate all images into three groups. Each group contains one type of MR image based on the image view. In each group, about 900 images are used for the training procedure and about 200 images are used as test images.

Our network uses binary cross-entropy as the loss function and the network is tuned using this parameter.

IV. EXPERIMENTAL RESULTS

Our network has been implemented on a server with Intel Core i7-4790K processor, 32 GB of RAM, and two NVIDIA GeForce GTX Titan X GPU cards with scalable link interface (SLI).

Comparing segmented images to evaluate the quality of segmentation is an essential part for measuring the progress the neural network. For comparison purposes, we used a simple network which has a probability map concatenation. The probability map is obtained from the ground truth binary maps of the training images to show probability of a pixel being tumor. We evaluate the segmentation results using Dice criterion. The Dice coefficient, also called the overlap index, is a metric for validation of medical image segmentation. The pair-wise overlap of the repeated segmentations is calculated using the DICE, which is defined by:

$$DICE = \frac{2TP}{2TP + FP + FN}$$

where $TP$ is true positive results or correctly segmented tumor pixels, $FP$ is false positive, and $FN$ is the false negative results of the segmentation. False positive results are obtained when a pixel which is not tumor is classified as tumor. Also, FN is referred to the number of pixels that are tumor and are falsely labeled as non-tumor. As shown in Table 1, the results for LinkNet networks that are trained individually by different angles are better than one LinkNet network that is trained by all dataset without separating. this shows the importance of detaching the dataset.

Table 1: Results of different approaches

| Method | Data | Dice |
|---|---|---|
| Single LinkNet for all directions | All angles | 0.73 |
| Separately trained Linknet networks for each direction | Coronal view | **0.78** |
|  | Sagittal view | **0.79** |
|  | Axial view | 0.71 |

V. CONCLUSION

In this paper, we introduced a new method for CNN to automatically segmenting the most common types of brain tumor, i.e. the Glioma, Meningioma, and Pituitary. This technique does not require preprocessing steps. The results show that the separation of images based on angles improves segmentation accuracy. The best Dice score that was obtained is 0.79. This relatively high score was obtained from segmentation of tumors in sagittal view images. Sagittal images do not contain details of other organs and tumor is more prominent than other images. The lowest Dice score in our experiments was 0.71 which is related to the images from the axial view of the head.

As compared to other images, axial view contains fewer details. It is expected that by performing preprocessing on this group of images better classification of tumor pixels could be obtained and the Dice score will increase.

Our proposed method may be implemented as a simple and useful tool for doctors in segmenting of brain tumor in MR images.